\documentclass[11pt]{article}
\usepackage{graphicx}
\usepackage{times}
\usepackage{latexsym}
\usepackage{amsmath}
\usepackage{multirow}
\usepackage{url}
\usepackage{acl2012}
\title{ Precision-biased Parsing and High-Quality Parse Selection }

\author{Yoav Goldberg \\
  Ben Gurion University of the Negev \\
  Department of Computer Science \\
  POB 653 Be'er Sheva, 84105, Israel \\
  {\tt yoav.goldberg@gmail.com} \\\And
  Michael Elhadad \\
  Ben Gurion University of the Negev \\
  Department of Computer Science \\
  POB 653 Be'er Sheva, 84105, Israel \\
  {\tt elhadad@cs.bgu.ac.il} \\}

\date{}

\begin{document}
\maketitle

\begin{abstract}
We introduce precision-biased parsing: a parsing task which favors precision over recall by allowing the parser to abstain from decisions deemed uncertain.  
We focus on dependency-parsing and present an ensemble method which is capable of assigning parents to 84\% of the text tokens while being over 96\% accurate on these tokens.
We use the precision-biased parsing task to solve the related high-quality parse-selection task: finding a subset of high-quality (accurate) trees in a large collection of parsed text.  We present a method for choosing over a third of the input trees while keeping unlabeled dependency parsing accuracy of 97\% on these trees.
We also present a method which is not based on an ensemble but rather on directly predicting the risk associated with individual parser decisions.  In addition to its efficiency, this method demonstrates that a parsing system can provide reasonable estimates of confidence in its predictions without relying on ensembles or aggregate corpus counts.
\end{abstract}

\section{Introduction and Methodology}

Parsing technology has made great progress over the last decade, and current state-of-the-art parsers for English have reported accuracies in the low 90\%'s.  
Current parsing systems are designed to provide a \textit{complete} parse to \textit{every} sentence, and are evaluated based on their average number of correct decisions over a test corpus.  Such evaluation, however, does not tell a complete story, as the mistakes are never uniformly distributed across sentences.  In practice, parsers usually perform very well on some sentences, but also perform very poorly on others.  In addition, parsers do not provide confidence estimates in their predictions, making it hard for downstream applications to rely on parsers' output, even if the average parsing quality is high.  

We advocate a modification to the parsing task, which we term \textit{precision-biased} parsing.  Rather than providing a complete parse to every sentence, we advocate providing \textit{partial analyses} to \textit{some} sentences, while trying to guarantee that the structures that are provided are of high quality.  In other words, we advocate parsing systems which are able to trade recall for precision.  The trade-off can occur either at the sentence level (abstaining from providing a parse to some sentences) or at the individual attachment level (abstaining from attaching some words or phrases to the rest of the sentence in case the attachment is uncertain).

Such trade-off is useful for tasks which rely on precise structures.\footnote{The other direction, of trading precision for recall, is also of interest.  A solution to this inverse problem is already available to some extent in the form of k-best parsing and packed forests.} 
These include information extraction and question answering systems, sentence simplification and summarization systems and syntax-based translation, as well as linguistically-oriented tasks such as learning selectional preferences, case frames or lexical ontologies.  In addition, partial but precise output may prove useful for self-training \cite{mcclosky2006self-training}, uptraining \cite{petrov2010uptraining} and active-learning setups.

Some previous efforts \cite{yates2006woodward,reichart2007sepa} attempt to identify high-quality parse trees among parser's output.  This is an instance of trading recall for precision at the sentence level.  Here, we focus on trading recall for precision at the individual attachment level.  As discussed in section \ref{sec:approach}, solving the problem at the individual attachment level entails a natural solution to the problem at the sentence level as well.  Moreover, we believe there is benefit in solving the problem at the attachment level -- useful information can be extracted from a partial parse tree, even when some attachment decisions are missing or marked as unreliable.

The precision-biased task was explored in the past in the context of parsers based on manually-developed grammars \cite{briscoe2002precision,briscoe2005gr}. However, state-of-the-art data-driven statistical parsers do not allow trading recall for precision.
In this paper, we focus on data-driven dependency parsing.

We begin by defining the precision-biased parsing task and its evaluation measures (Section \ref{sec:define}), and provide a strong baseline based on parse-ensembles (Section \ref{sec:baseline}).  While effective, the ensemble system requires substantial computational effort, and is of little theoretical interest as it is well known that committee (dis)agreement is a good indicator of confidence.   We propose another method based on parser-modeling in Section \ref{sec:system}.  We train a probabilistic classifier to try and predict the risk associated with attachment decisions in the parser's output.  The classifier learns the error patterns of the parser, and assigns a reliability score to parse edges.  By thresholding these reliability scores, we can effectively trade recall for precision.  The method comes close to the ensemble-based baseline in terms of precision and coverage, while running faster and providing a straightforward way to control the recall/precision tradeoff.  More importantly, it demonstrates that reasonable confidence estimates of the correctness of parser predictions can be attained without relying on committee, diversity, or aggregate corpus-based counts of recurring structures.  Inspecting the behavior of the learned model on PP-attachment, a classic case of syntactic ambiguity, reveal that it is not judged by the model as being categorically hard. Rather, some PP-attachment cases are marked as unreliable, while others are not.

\section{Precision-biased Dependency Parsing}
\label{sec:define}

In the traditional dependency parsing task, the input is a set of sentences (the test corpus), and we are interested in the most precise analysis for each sentence.  The task performance is measured based on the average number of tokens that got assigned a correct parent over the entire test corpus.  Crucially, the parsing process in this task must assign a parent to each of the tokens in the test corpus.  

In contrast, in the \textit{precision-biased} task we are concerned with precision more than recall.  We allow the parsing system to \textit{abstain} from providing an analysis to some of the input by \textit{skipping} some decisions. That is, we require the parser to assign parents to as many of the input tokens as possible, but allow it to leave the parents of some tokens unassigned.  

\noindent\textbf{Metrics} Let $T$ be the set of input tokens, $A$ be the set of tokens that got assigned a parent, and $S$ be the set of tokens that were not assigned a parent ($T=A\cup S$, $A\cap S=\emptyset$).  Let $C\subset A \subset T$ be the set of tokens that got assigned a correct parent.  Then:
\begin{align*}
{precision} = \frac{|C|}{|A|} & \hfill & 
{recall}    = \frac{|C|}{|T|} & \hfill &
{coverage}  = \frac{|A|}{|T|} & &
\end{align*}

By requiring complete coverage of the input tokens (${coverage}=\frac{|A|}{|T|}=1$) we get $A=T$, and then ${precision}={recall}={accuracy}$, where \textit{accuracy} is the traditional dependency parsing accuracy.

The precision-biased setting allows \textit{coverage}~$<1$.  The aim is to maximize \textit{precision}, while still retaining sufficient \textit{coverage} of the input tokens.\footnote{We chose to balance \textit{precision} against \textit{coverage} rather than against \textit{recall} because \textit{coverage} is an upper-bound on \textit{recall}: if \textit{precision}$=1$ then \textit{coverage}$=$\textit{recall}.}

\vspace{5pt}

\noindent\textbf{Parse-selection} Previous work addresses the \textit{parse-selection} task: selecting a subset of the input sentences for which we have high-accuracy parses.  This is an instance of precision-biased parsing in which abstaining on a token requires abstaining also on all the other tokens in the same sentence.  
When discussing the parse selection task we distinguish between \textit{token coverage} which is identical to coverage as defined above, and \textit{sentence coverage} which is the number of selected sentences divided by the total number of input sentences $\frac{|\{sent|sent\subset A\}|}{|\{sent|sent\subset T\}|}$.  The definitions of precision and recall remain as above.

\section{Our Approach}
\label{sec:approach}

We tackle precision-biased parsing by defining a \textit{riskiness} function on individual attachment decisions.  The riskiness $R(tok,par)$ of a token/parent pair is the inverse of our confidence in the attachment decision.  A high riskiness indicates uncertainty in the decision, and low riskiness indicates that we believe the decision to be correct.  We then set a riskiness-threshold, and abstain from any parent assignment for which the riskiness is above the threshold.  Setting a low riskiness-threshold results in higher precision (considering even low-risk decisions as too risky), and setting a high riskiness-threshold results in higher coverage.

\paragraph{Parse-selection} Having defined a risk function and a risk-threshold, we get a natural selection criteria for the parse-selection task: high-quality parses are those for which at most $K$ attachments are above the riskiness threshold.  The precision/coverage balance can be controlled either by changing the risk threshold, or by changing $K$.  

\subsection{Related Work}

While little research attention was dedicated to the precision-biased task \cite{briscoe2002precision,briscoe2005gr}, several studies address the parse-selection task.  
Yates et.al. \shortcite{yates2006woodward} perform parse-selection by filtering out parses containing ``semantically implausible'' relations, where semantic-plausibility is estimated by high co-occurrence of the words in relation in a large corpora ({\em i.e.}, the web).

Reichart and Rappoport \shortcite{reichart2007sepa} perform committee based selection of high-quality constituency-parses by calculating an agreement measure between 20 copies of a lexicalized parser trained on different subsets of a training corpus.

Sagae and Tsujii \shortcite{sagae2007adaptation} select high-quality dependency parses by using two dependency parsers and selecting only sentences on which both parses agree on the entire parse.

Kawahara and Uchimoto \shortcite{kawahara2008reliable} identify high-quality dependency parses by training a classifier based on sentence level features: sentence length, average dependency length, number of unknown words, number of commas and conjunctions, and corpus frequencies of sentence words.

Reichart and Rappoport \shortcite{reichart2009pupa} identify high-quality parses of an unsupervised parser by looking for parses with many reliable constituents, where reliability of a constituent is calculated based on the number of times its POS-sequence appears in the automatically parsed text. Finally, Dell'Orletta \textit{et.al,} \shortcite{ulissee2011} assign quality-scores to dependency-parses using a metric which measures various syntactic properties of the parse tree and compares them the aggregate measurements over the entire parsed corpora.

To summarize, there are three lines of work addressing the parse-selection task: selecting parses based upon agreement between a committee of parsers, selecting parses based on agreement between the parses and aggregate counts over a large corpora (either lexicalized ``semantic'' agreement or syntactic agreement), and selecting parses based on sentence-specific features (length, vocabulary, number of commas and so on).  

In contrast, we are primarily interested in selecting high-quality edges rather than complete parses. We view parse-selection as an extension of the precision-biased parsing task, and perform parse-selection based on the number of risky attachment decisions. Our assessment of the riskiness or reliability of a particular decision is not based on aggregate corpus counts nor on global features of the input sentence (though such kinds of information may be integrated in the future).  In our first method, we adopt a committee-based approach, but apply it primarily for edge-selection. In the second method we present below, we investigate features which may help the parser assess edge riskiness.  We note that the marginal edge probabilities obtained from a log-linear parsing model as in \cite{smith2007probmtt} are not reliable predictors of edge riskiness: indeed, the pruning procedure used in \cite{collins2008tag} consider edges with marginal scores of up to $10^{-6}$ of the highest scoring edge as possible candidates in order to ensure sufficient coverage, indicating that such models may greatly overestimate the marginals of incorrect edges, while underestimating the marginal values of correct edges.

Kawahara \shortcite{kawahara2001case-frames} present an automatic method for Case-Frame dictionary construction for Japanese.  Their method identify verb case-frames by identifying reliable syntactic constructions, where the reliability is learned using a hand-crafted heuristic and aggregate corpus counts.  This demonstrates the usefulness of identifying reliable instances of specific constructions.  Our proposal is to try and identify reliable instances of many different constructions, without relying on hand-crafted heuristics.

\section{Data}
\label{sec:data}

Our experiments are based on the dependency-version of the Penn WSJ corpus, as converted using the Penn2Malt\footnote{http://http://w3.msi.vxu.se/$\sim$nivre/research/Penn2Malt.html} software with Collins' head-rules.
The data is POS-tagged using the HMM-based Hunpos tagger\footnote{http://code.google.com/p/hunpos/}.

While our work is not directly comparable to any previous work\footnote{We are not aware of previous work on the precision-biased task, while previous work for the parse-selection task either focus on constituency-structures, or use non-standard datasets.}, this presents an opportunity to stop following the standard train/test/dev splits, and in particular to stop testing only on section 23.  Instead, we adopt a setup in which we use sections 2-11 (about 18k sentences) for training the parser(s), sections 12-15 (8900 sentences) for training the riskiness-estimator (where appropriate), section 16 (2780 sentences) for development and sections 17-21 (9500 sentences) for testing.  This setup leaves a reasonable amount of training data for the statistical models (the parser training set is roughly the same size as the one used in the CoNLL shared task), while retaining a much larger test set than the standard setting.

\section{Parser-ensemble Riskiness Estimation}
\label{sec:baseline}

Our first method of estimating the riskiness function is using an ensemble method.  Ensemble methods have been shown to provide good results for dependency parsing \cite{sagae2006blend,malt-blended}, as well as for parse-selection as discussed above.  Here, we use ensembles to estimate the riskiness of individual edges in a dependency tree.  To estimate the riskiness, we parse the input sentence using $k$ different parsers and take the intersection of their predictions.  The riskiness of a token/parent pair is 0 if all $k$ parsers agree
 on that prediction, and 1 otherwise.  In the final output, we take only edges with 0 riskiness.

 We use an ensemble of 3 parsers: a linear-time shift-reduce parser as described in \cite{wenbin-arcstandard} (\textsc{ShiftReduce}), the globally optimized first-order projective dependency parser of \cite{mst} (\textsc{Mst1}), and the easy-first parser of \cite{naacl2010bidi} (\textsc{EasyFirst}).  Such ensemble was shown in \cite{naacl2010bidi} to provide good oracle accuracies, as well as state-of-the-art accuracies in a non-oracle setting due to the diversity among its parsers.  The runtime of this ensemble is dominated by the $O(n^2)$ feature extraction stage and the $O(n^3)$ inference of the globally optimized \textsc{Mst1} parser.  

\subsection{Results and Discussion}

The individual parser's scores on the test set are 87.4 (\textsc{ShiftReduce}) 88.6 (\textsc{Mst1}) and 88.4 (\textsc{EasyFirst}).

\noindent\textbf{Precision-biased Scores} The precision-biased scores of the ensemble system on the test-set are 96\% precision with a coverage of 84.2\% (recall of 80.8\%).  By not providing an analysis for about 15\% of the input tokens, we get an impressive gain in precision.  

\noindent\textbf{Parse-selection Scores} As discussed above, we reduce parse-selection to risk-based precision-biased parsing by selecting parses with at most $K$ risky attachments.  Table \ref{tbl:ensemble-parse-selection} shows the precision and sentence-coverage on the development and test set for various values of $K$.  With a $K$ value of 0 (forcing the parsers to agree on all edges) achieves a precision of 97.5 while covering about a quarter of the sentences in the test set.  By allowing one disagreement between the parsers the precision drops to 95.0, but we gain a better sentence-coverage -- about 36\%.  Increasing the value of $K$ decreases the selected parses accuracy while increasing their quantity.

\begin{table}[h]
   \centering
   \scalebox{0.9}{
   \begin{tabular}{c||c|c}
      \textbf{K} & \textbf{Precision (\%)} &  \textbf{Sentence-Coverage (\%)} \\
        & dev / test &   dev / test  \\
        \hline
      0 & 97.5 / 97.8 & 23.2 / 24.6 \\
      1 & 95.0 / 95.6 & 36.2 / 36.8 \\ 
      2 & 93.0 / 93.4 & 47.2 / 47.3 \\
      3 & 91.0 / 91.3 & 56.9 / 57.1 \\
      4 & 89.4 / 89.5 & 64.5 / 66.0 \\
   \end{tabular}}
   \label{tbl:ensemble-parse-selection}
   \caption{\footnotesize{Ensemble-based Parse-Selection Precision and Coverage for various risk-cutoffs ($K$)}}
\end{table}

\section{Single-parser Riskiness Estimation}
\label{sec:system}

While the ensemble method is effective at the precision-biased task, it has two shortcomings: (1) it takes a long time to run due to the runtime complexity of the \textsc{Mst1} parser, and (2) it does not provide a way of tuning the precision/coverage balance.

Here we take a different route.  We use a single parser (we use the \textsc{EasyFirst} parser for its balance between speed and accuracy and its incremental parse construction), and train a discriminative probabilistic classifier to predict the risk associated with its predictions.  

\textsc{EasyFirst} is a greedy parser that work by incrementally adding dependency edges in a bottom-up fashion (see \cite{naacl2010bidi} for the details).
It 
is trained to take easy decisions before harder ones, but does not provide confidence in its predictions: at a given step, the highest scoring action can still be very ambiguous, yet easier than the alternatives.  
The two measures of the best possible (``easiest'') action and the riskiness of an action are interrelated, but not identical.
The easiest action, at a given stage, may still be risky.  
For example, consider the case in which the parser sees a configuration consisting of [Verb Noun Prep].  This is a PP attachment ambiguity, where the Prep should be the child of either the Noun or the Verb.  Concretely, the parser should choose to either attach Noun under Verb and then Prep under Verb+Noun, or to first attach Prep under Noun and then Noun+Prep under Verb.  At this stage, the two possible attachments (Verb+Noun and Noun+Prep) are risky (even though the Verb+Noun edge will turn out in the final parse in any case), but the parser should nevertheless choose one of them.  It will choose, based on its training experience and on the specific properties of the VP, NP and PP at hand, the action which it finds is most correct.  This would be the least-risky attachment \emph{at a given stage}, but it does not reflect directly on the objective riskiness of the decision at large.  

In the precision-biased setting, we are interested in assessing the objective riskiness of various parser decisions. 

\noindent\textbf{Riskiness Predictor}
We train a separate classifier to assess the riskiness involved in each prediction.  We interpret the riskiness as a probability function: $risk(\textit{context})=Pr(\textit{decision is wrong}|{context})$.  In words: the riskiness is the probability of the parser making a wrong choice in a given situation (context).  The riskiness function does not necessarily depend on the actual decision ({\em i.e.}, it should be interpreted as ``when faced with situation X you are likely to make a mistake'' rather than as ``attaching $t_i$ below $t_j$ in situation X is likely to be wrong''), but the decision can be encoded in the context if desired.

We treat riskiness prediction as a binary classification task, and fit a Maximum Entropy model\footnote{We use the Megam optimizer\cite{megam}\footnote{http://www.cs.utah.edu/$\sim$hal/megam}} based on training data as described in Section 7.1 below.

We experimented with several alternative interpretations of the riskiness function, capturing different kinds of information (features).

\vspace{5pt}

\noindent\textbf{Riskiness of parser actions}
An interesting question that arises is whether the information available during parsing is sufficient for determining the risk associated with a parsing decision, and which kinds of information are most useful.

The first set of experiments attaches risk to \textit{parser actions}.  These aim to answer the question ``can the parser assess the quality of its own actions''.
Note that parser actions are not equivalent to attachment decisions: the easy-first parser may choose to attach a token to its correct parent and still be wrong, because it is not yet the correct time to do so (because the child node is not yet saturated), and this action, while resulting in one correct edge, prevents future correct edges from being added (consider the Verb+Noun edge in the PP-attachment example above).  Thus, this set of experiments can be used only for the parse-selection task (selecting as good parses those for which there were less than $K$ risky actions), and not for the precision-biased parsing task.  

We experimented with the following feature sets:

\noindent\textit{Process-based features}: \texttt{action\_process} is a minimal set of 5 numerical features which relate only to the parsing process itself.  These include: sentence-length, current number of parent-less tokens, score of the best action (to be applied), score of the second best action, and the difference between the best and second-best actions.

\noindent\textit{State-based features}: \texttt{action\_state} is a set of features which relate only to what the parser sees.  Here, we use the exact same feature set which is used by the parser for predicting the scores of the various actions. 

\noindent\textbf{Riskiness of predicted edges}
The second set of experiments associates riskiness with edge predictions.  That is, riskiness is interpreted as ``what is the probability of this particular predicted edge to be wrong''.  In contrast to the previous experiments, this definition of riskiness addresses the full precision-biased parsing problem, by abstaining from providing (or ruling-out) attachments for edges that are considered too risky.  

We experimented with the following feature sets:
\noindent\textit{State-based features}: as above, the \texttt{edge\_state} feature set encodes exactly what the parser sees when making an attachment decision, {\em i.e.}, the feature set used by the parser when scoring actions.  However, here wrong parser actions which result in a correct edge are considered as correct (non-risky) examples.

\noindent\textit{Edge-factored features}: the \texttt{edge\_factored} feature set is not related to the parsing process, and can be extracted from the parse tree in a post-processing step.  Here, we use the same features as used in Ryan McDonald's first-order edge-factored MST parser \cite{mst}.

\noindent\textit{Higher-order features}: the \texttt{edge\_higher} feature-set does not depend on the parsing process, and uses more information than the edge-factored one: the features of a (token,parent) pair include information on the token and the parent, as well as on the siblings of the token, siblings of the parent, children of the token, and parent of the parent. 

\texttt{edge\_state} has only a negligible effect on the parsing time (as above, the features are already extracted by the parser), while \texttt{edge\_factored} and \texttt{edge\_higher} have a noticeable (though still small) effect on the parsing time by adding $n$ feature extraction and scoring operations.  

\section{Experiments and Results}

\subsection{Training} We followed the following procedure:

\noindent1. Train the easy-first parser on the parser-training set, and use it to parse the rest of the data (riskiness-training, test and dev, see Sec. \ref{sec:data}) while keeping track of the parser's predictions.\footnote{In case of a smaller treebank, a k-fold jacknifing scheme should be used.} 

\noindent2. Extract correct and incorrect decisions and their corresponding features (according to the definitions above) from the automatically parsed data.

\noindent3. Train a MaxEnt binary classifier on the riskiness-training set.

\subsection{Evaluation}

\begin{figure}[t!]
   \includegraphics[width=0.5\textwidth]{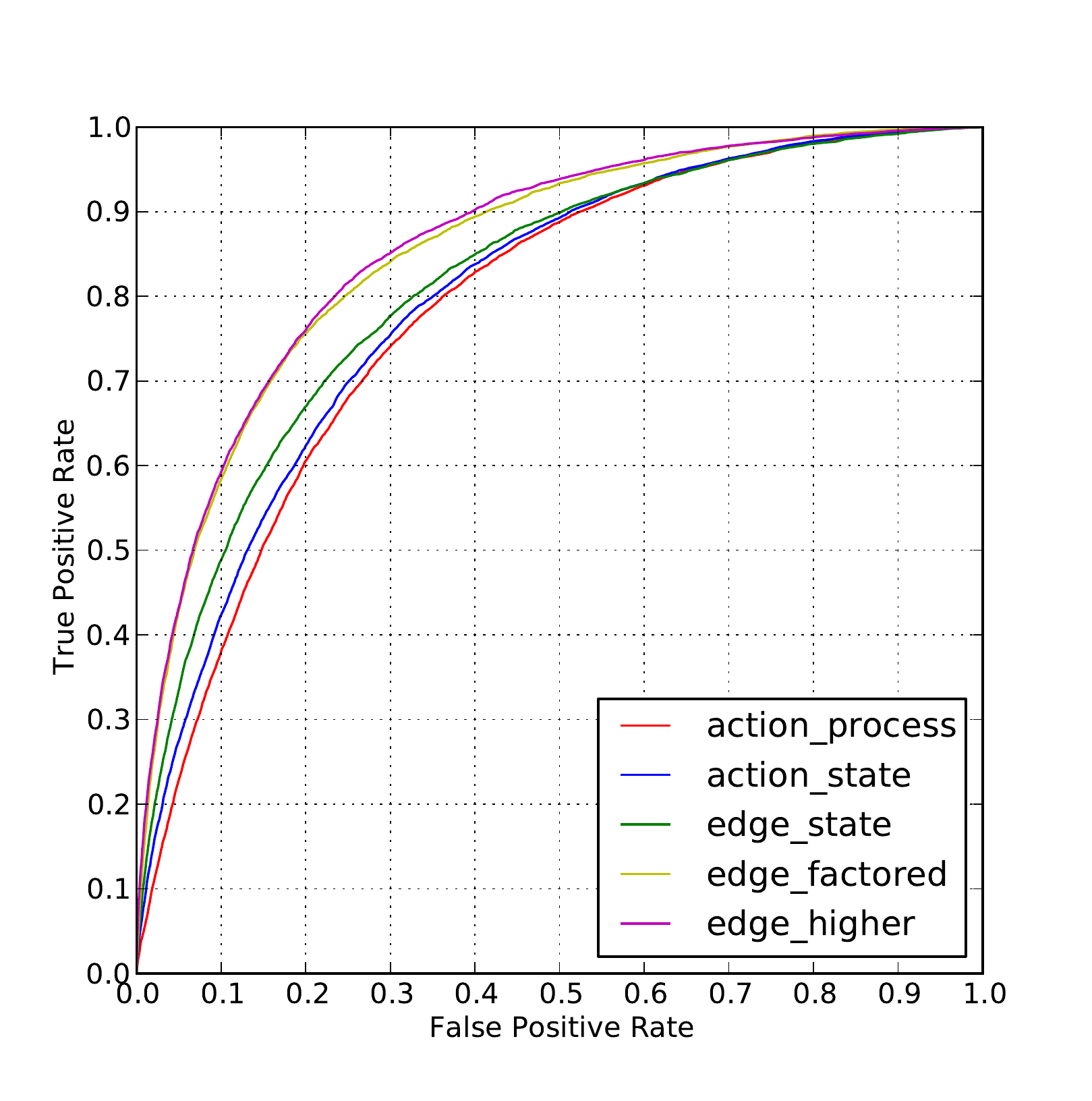}
   \caption{\footnotesize{ROC Curves for the various riskiness predictors on the dev set.  The True Positive Rate is the percentage of incorrect decisions which were identified as risky, and the False Positive Rate is the percentage of correct decisions which were identified as risky.}}
   \label{fig:roc}
\end{figure}

\noindent\textbf{ROC}
We begin by plotting the ROC curves for identifying risky decisions using the different MaxEnt risk predictors with varying risk-thresholds.  Figure \ref{fig:roc} presents the results.  The curves are not entirely comparable: the two lower curves (\texttt{action\_process} and \texttt{action\_state}) identify risky parser actions, while the higher curves identify risky edge attachments.  
There is a clear hierarchy between the different predictors, but even the simplest ones are quite effective at identifying risky decisions.
The predictors that attach riskiness to edges are more effective than those that attach riskiness to parsing actions, even when the same feature-set is used (\texttt{edge\_state} vs. \texttt{action\_state}), and the two predictors that use information external to the parser (\texttt{edge\_factored} and \texttt{edge\_higher}) are better than those using information internal to the parser.  
Still, it is interesting to note that the exact same feature-set which is available to the parser during parsing is sufficient to assess the riskiness of many of the decisions which are based on the same feature-set.

There is only a small difference between the \texttt{edge\_factored} feature-set and the one including higher-order features (\texttt{edge\_higher}): while the extra contextual information does help, most of the riskiness associated with an edge can already be determined based on the edge itself and sentence-level properties (without considering proposed surrounding edges).

\begin{figure*}[pth!]
   \includegraphics[width=\textwidth]{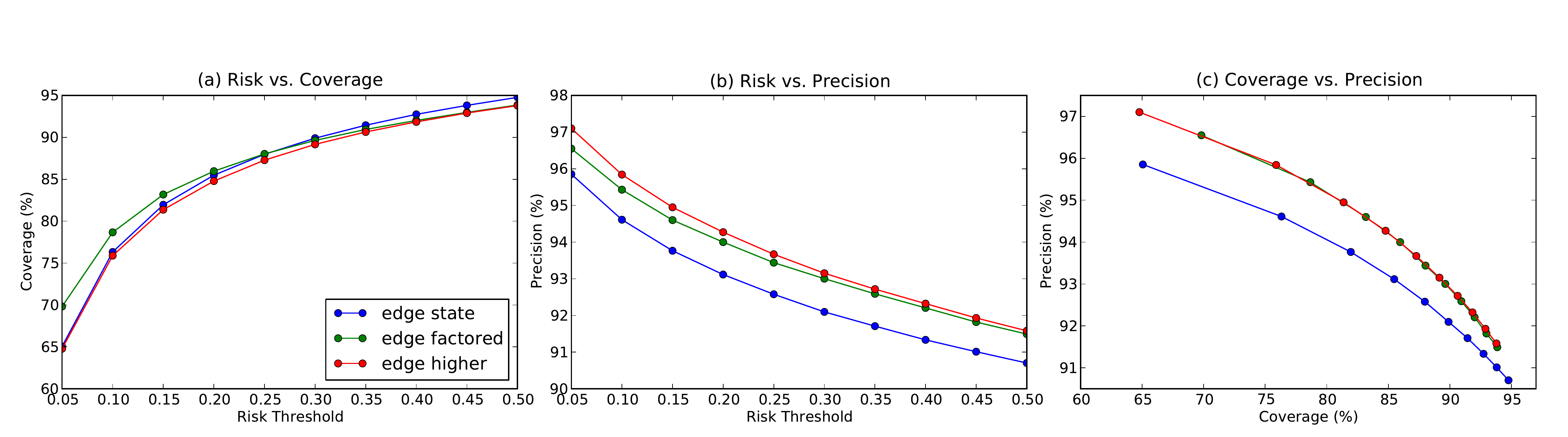}
   \caption{\footnotesize{Precision-biased results on the dev-set.}}
   \label{fig:prec-based-curves}
\end{figure*}

\noindent\textbf{Precision-biased Scores} We now turn to evaluating the precision-biased results for the various predictors.  As discussed above, these can be calculated only for the three edge-based riskiness predictors.  For the precision-biased results, the parser abstains from predicting edges with an associated riskiness above a certain riskiness-threshold.  Figure \ref{fig:prec-based-curves} plots the precision and coverage of the parser for varying riskiness-thresholds, using the three different riskiness-predictors.  The third plot in the figure plots precision against coverage for the same predictors.  The overall trends are similar to those observed in the ROC curves, though here the difference between \texttt{edge\_factored} and \texttt{edge\_higher} is somewhat more pronounced.  

With appropriate riskiness thresholds we could achieve a coverage as high as 95\%, or precision of above 97\%.  Unfortunately, we cannot get both: higher precisions mean lower coverages and vice-versa.  
Compared to the ensemble-based riskiness estimation (96\% precision with 84\% coverage) the single-parser results are not as strong.  The same level of coverage (84.6\%) results in a precision of around 94.2\%, and a precision of 96.5\% leads to a coverage of 70\%.  A riskiness-threshold of 0.15 (\texttt{edge\_higher} predictor) strikes a nice balance of just over 95\% precision with 80\% coverage.  A coverage of 90\% gets us a precision of above 92\%, still substantially higher than the 88.4\% of the full-coverage baseline parser.
The numbers are practically the same for the development and test sets.

\begin{figure}[pth!]
   \includegraphics[width=0.5\textwidth]{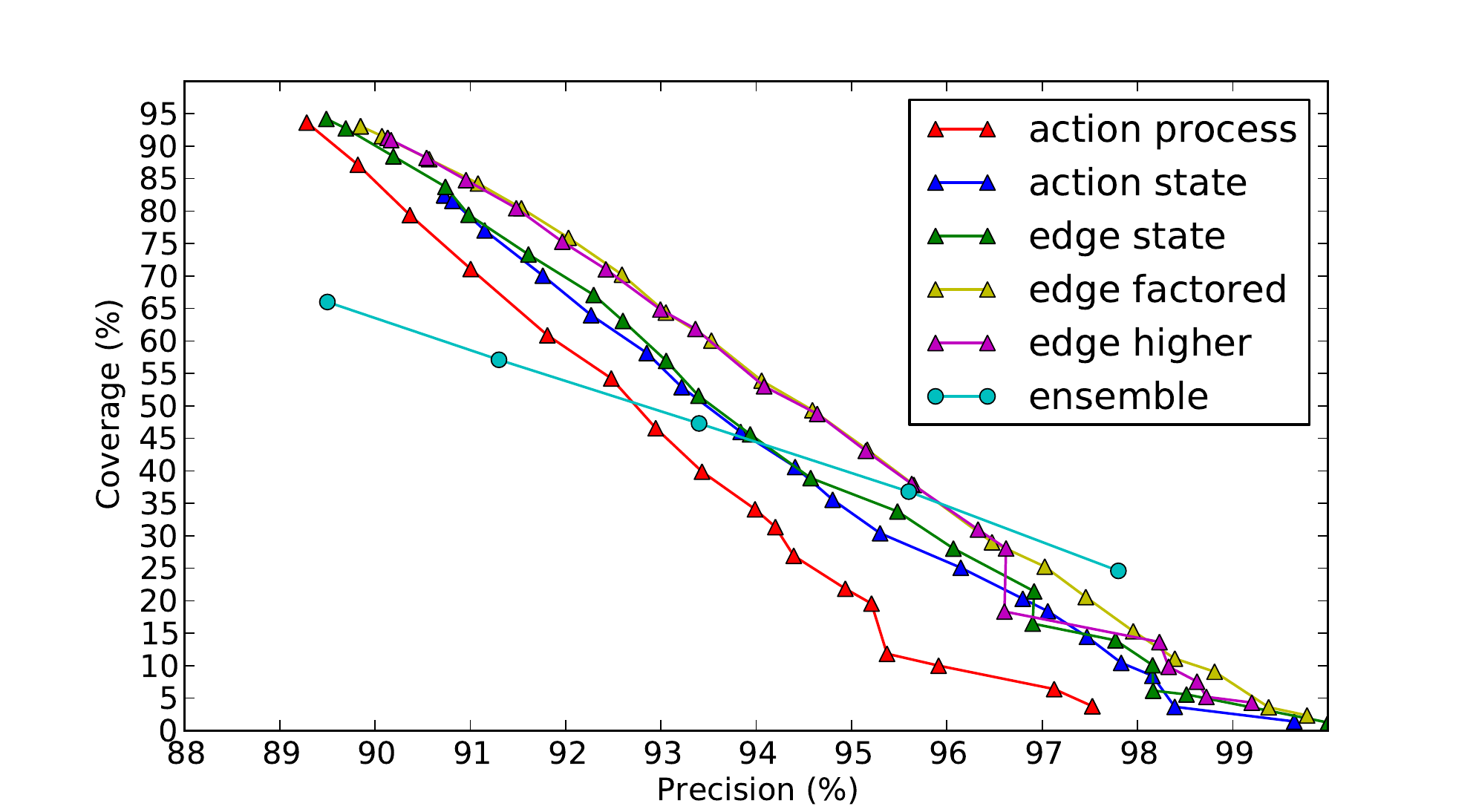}
   \caption{\footnotesize Parse-selection (precision vs. sentence-coverage) results on the test sets based on best ($K$,$R$) values obtained on dev set.}
   \label{fig:risksel}
\end{figure}

\noindent\textbf{Parse-selection scores} 
The parse-selection task has two tunable parameters: the riskiness-threshold $R$, and the number of risky decisions ($K$) above which we regard the complete parse-tree as unreliable.  For each predictor we performed a grid-search over these parameters with $R$ ranging in value from 0 to 0.5 with increments of 0.01, and $K$ ranging from 0 to 4.  For each point we recorded the precision and the sentence-coverage over the development set.  We then chose the parameters yielding the best sentence-coverage for a given precision level (we varied the precision levels from 89 to 99 with increments of 0.5).  We processed the test-set with the selected parameter-values.
Figure \ref{fig:risksel} plots precision against sentence-coverage on the test-set using the ($K$,$R$) obtained on the development set.

\section{Riskiness of PP-attachments}
Having developed a model of assessing the riskiness associated with parsing decisions (i.e., the chances of a certain decision being wrong), what does it find to be risky? A complete analysis of the behaviour is well beyond the scope of this paper, but we provide a glimpse into what is possible by inspecting the model's behaviour on a particular case of syntactic ambiguity: PP attachment.  Not surprisingly, attaching a preposition to its parent is judged very risky.  When sorting POS-tags by the number of times their parent-attachment is judged to be risky, prepositions are at the top of the list.  But do the models learn that all PP attachment decisions are risky and abstain from attaching any PP to their parent? Or maybe some kinds of prepositions riskier than others?  Table \ref{tbl:pp1} shows the confusion matrix for prepositions as judged by the \texttt{edge\_higher} model with a risk-threshold of 0.15.  Table \ref{tbl:pp2} breaks down the numbers by preposition type.

\begin{table}[th]
\centering\scalebox{0.9}{
   \begin{tabular}{c|cc}
   reality / model & Risky & Safe \\
   \hline
   Incorrect       &  TP: 961  &   FN: 326    \\
   Correct         &  FP: 1353     & FP: 4302     \\
   \end{tabular}}
   \caption{\footnotesize{Preposition's riskiness confusion matrix over dev-set.  \texttt{edge\_higher} features, riskiness-threshold of 0.15.}}
   \label{tbl:pp1}
\end{table}
\begin{table}[th]
\centering\scalebox{0.9}{
   \begin{tabular}{c||cccc|c}
   Preposition & TP & FP & TN & FN & Total\\
   \hline
as & 41 & 80 & 123 & 19 & 263   \\
with & 34 & 76 & 168 & 12 & 290   \\
at & 55 & 68 & 177 & 21 & 321   \\
on & 77 & 90 & 179 & 16 & 362   \\
from & 32 & 59 & 222 & 17 & 330   \\
that & 31 & 62 & 223 & 16 & 332   \\
by & 35 & 49 & 261 & 9 & 354   \\
for & 120 & 186 & 317 & 28 & 651   \\
in & 245 & 319 & 575 & 75 & 1214   \\
of & 18 & 40 & 1457 & 39 & 1554   \\
   \end{tabular}}
   \caption{\footnotesize{Preposition's riskiness by type over dev-set. \texttt{edge\_higher} features, riskiness-threshold of 0.15. TP: risky/incorrect, FP:risky/correct, TN:safe/correct, FN:safe/incorrect}}
   \label{tbl:pp2}
\end{table}

Interestingly, while PP attachment is the most risky phenomena, most PP attachment cases are correctly judged by the model to be non-risky (4302 cases).  326 other PP attachment cases are judged by the model to be safe, but are incorrect.  Finally, the model marks 2314 PP-attachment cases as risky, and 961 of these are indeed parsing mistakes.  

When breaking down by preposition type, we can observe that ``of'' is by far the least ambiguous (1457 cases, or 93\%, are correctly marked as safe), ``in'',``for'',``on'',``as'',``at'' are the most ambiguous (marked as risky about half of the time) while ``by'',``that'',``from'' are in between (25-35\% of the cases are judged to be risky).

\section{Discussion}
We advocate a modified version of the parsing task -- precision-biased parsing -- which favors precision over recall by allowing the parser to abstain from decisions about which it is uncertain.   In our view, partial but highly accurate structural information is in many cases more valuable than complete but less accurate structural information.  The precision-biased parsing problem is related to confidence estimation, that is, attaching reliability scores to model predictions.

In order to address the precision-biased parsing task we introduce the notion of riskiness of parser decisions.  On the basis of riskiness assessment, the parser can abstain from risky predictions.  This gives rise to a natural solution to the parse-selection task: reliable parse-trees are those associated with few risky actions.

After verifying that disagreement in a parser-ensemble is a good indicator of risky edges, we presented a novel approach that does not rely on a parser-ensemble, but instead learns to predict the riskiness involved with individual actions of a single parser.  While the method sacrifices more coverage than the parser-ensemble in order to achieve the same level of accuracy, the results are encouraging and demonstrate that a single parsing system can monitor the confidence of its own predictions.  Single parser riskiness assessment turns out to be a good indicator of confidence on aggregate: the single-parser system is as capable as the ensemble-based one at selecting high-quality complete parses.

\bibliography{../references}

\pagebreak

\begin{figure*}[h!]
   
   \noindent\textbf{\large{Supplementary Material}}


   \texttt{edge\_state}\footnotesize{, risk threshold 0.15:}
   \vspace{5pt}

   \includegraphics[width=\textwidth]{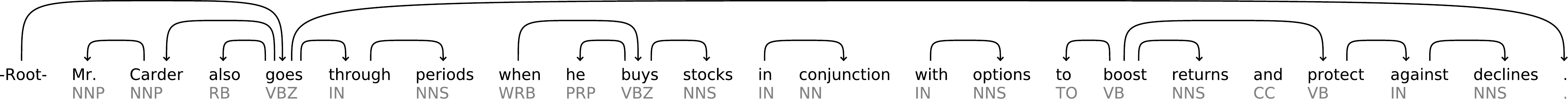}
   \vspace{20pt}
   \includegraphics[width=\textwidth]{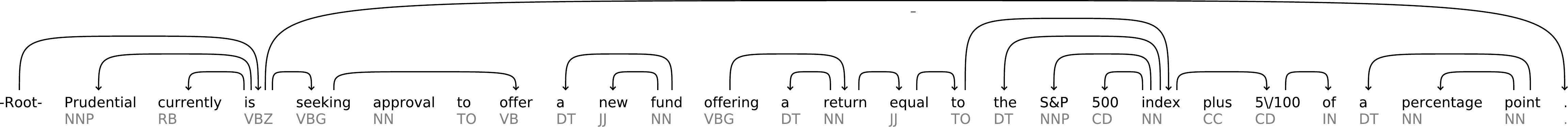}

   \texttt{edge\_factored}\footnotesize{,risk threshold 0.15:}

   \vspace{10pt}
   \includegraphics[width=\textwidth]{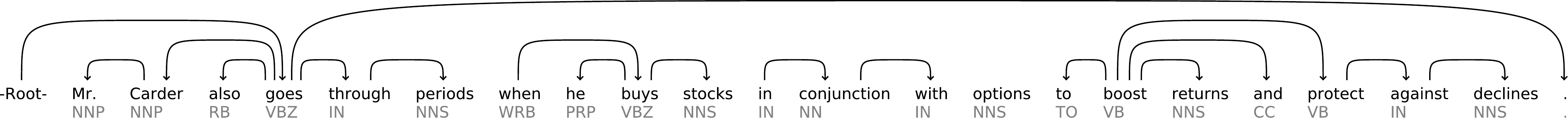}
   \vspace{20pt}
   \includegraphics[width=\textwidth]{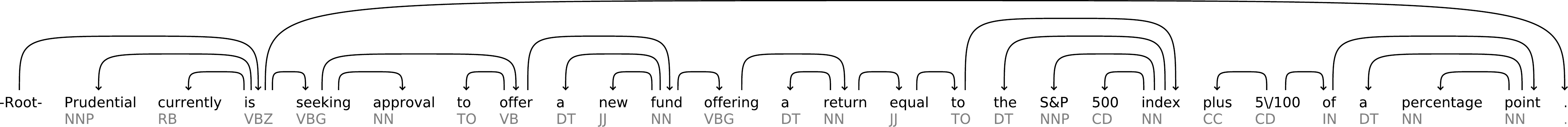}

   \texttt{edge\_higher}\footnotesize{, risk threshold 0.15:}

   \vspace{10pt}
   \includegraphics[width=\textwidth]{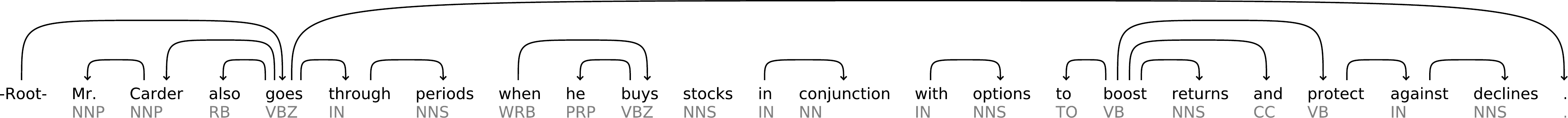}
   \vspace{20pt}
   \includegraphics[width=\textwidth]{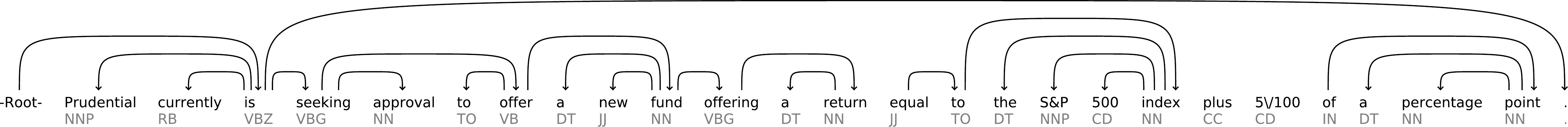}

   \caption{\footnotesize{Precision-biased parse examples of the single-parser systems' predictions on the dev set.}}
   \label{fig:partial-examples-edge-risk}
\vspace{-20pt}

\end{figure*}
\begin{figure*}[h!]
   \includegraphics[width=0.9\textwidth]{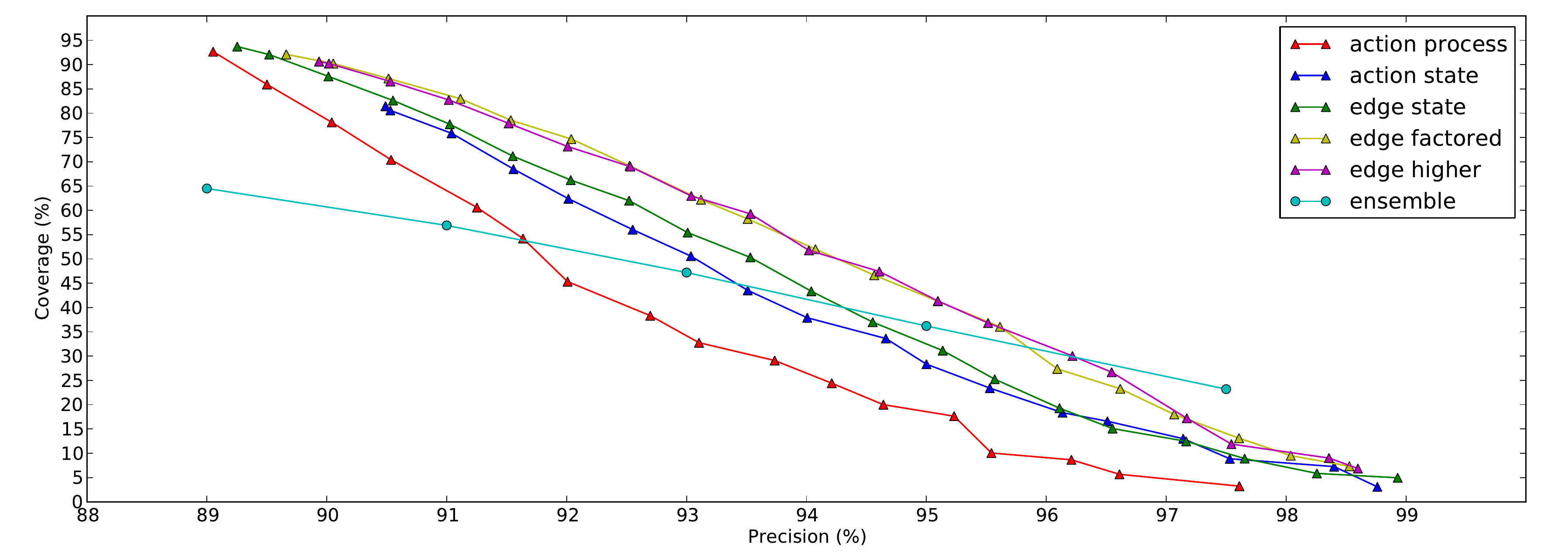}
   \caption{\footnotesize{Parse-selection based on single-parser risk predictors results (precision vs. sentence-coverage) on dev set.}}
   \label{fig:risksel-dev}
\end{figure*}

\end{document}